\documentclass[journal]{IEEEtran}

\usepackage{cite}

\usepackage{graphicx}
\usepackage{caption}
\usepackage[marginal]{footmisc}

\ifCLASSINFOpdf
\else
\fi
\usepackage{amsmath}
\usepackage{array}
\usepackage{booktabs}
\usepackage{algorithm}
\usepackage{algorithmic}
\usepackage{soul}
\soulregister\cite7
\soulregister\citep7
\soulregister\citet7
\soulregister\ref7
\soulregister\pageref7

\usepackage{xcolor}

\begin{document}

%
\title{Overcoming Long-term  Catastrophic Forgetting through Adversarial Neural Pruning and Synaptic Consolidation}

%
%
%

\author{Jian Peng, Bo Tang, Hao Jiang, Zhuo Li, Yinjie Lei, Tao Lin, and~Haifeng~Li*,~\IEEEmembership{Member,~IEEE}
\thanks{
This work was supported by the National Natural Science Foundation of China (Grant 41871364, 41861048, 41871302, and 41871276). This work was carried out in part using computing resources at the High Performance Computing Platform of Central South University.}
\IEEEcompsocitemizethanks{\IEEEcompsocthanksitem J. Peng, H. Jiang, Z. Li, and H. Li are with School of Geosciences and Info-Physics, Central South University, Changsha 410083, China.\protect\\
Corresponding author: H.Li lihaifeng@csu.edu.cn
\IEEEcompsocthanksitem B. Tang is with  Department of Electrical and Computer Engineering, Mississippi State University, MS 39762, USA.
\IEEEcompsocthanksitem Y. Lei is with college of electronics and information engineering, Sichuan University, Chengdu, China.
\IEEEcompsocthanksitem T. Lin is with College of Biosystems Engineering and Food Science, Zhejiang University, Hangzhou, China.}
\thanks{ J. Peng, B. Tang, H. Jiang, Z. Li, Y. Lei, T. Lin, H. Li. Overcoming Long-term Catastrophic Forgetting through Adversarial Neural Pruning and Synaptic Consolidation. IEEE Transactions on Neural Networks and Learning Systems. 2021. 10.1109/TNNLS.2021.3056201. \protect \\ }
}
%
%


\markboth{IEEE Transactions on Neural Networks and Learning Systems}
{Shell \MakeLowercase{\textit{Peng et al.}}: }
%



\maketitle

\begin{abstract}
Enabling a neural network to sequentially learn multiple tasks is of great significance for expanding the applicability of neural networks in real-world applications. However, artificial neural networks face the well-known problem of catastrophic forgetting. What's worse, the degradation of previously learned skills becomes more severe as the task sequence increases, known as the long-term catastrophic forgetting. It is due to two facts: first, as the model learns more tasks, the intersection of the low-error parameter subspace satisfying for these tasks becomes smaller or even does not exist; second, when the model learns a new task, the cumulative error keeps increasing as the model tries to protect the parameter configuration of previous tasks from interference. Inspired by the memory consolidation mechanism in mammalian brains with synaptic plasticity, we propose a confrontation mechanism in which Adversarial Neural Pruning and synaptic Consolidation (ANPyC) is used to overcome the long-term catastrophic forgetting issue. The neural pruning acts as long-term depression to prune task-irrelevant parameters, while the novel synaptic consolidation acts as long-term potentiation to strengthen task-relevant parameters. During the training, this confrontation achieves a balance in that only crucial parameters remain, and non-significant parameters are freed to learn subsequent tasks. ANPyC avoids forgetting important information and makes the model efficient to learn a large number of tasks. Specifically, the neural pruning iteratively relaxes the current task's parameter conditions to expand the common parameter subspace of the task; the synaptic consolidation strategy, which consists of a structure-aware parameter-importance measurement and an element-wise parameter updating strategy, decreases the cumulative error when learning new tasks.  Our approach encourages the synapse to be sparse and polarized, which enables long-term learning and memory. ANPyC exhibits effectiveness and generalization on both image classification and generation tasks with multiple layer perceptron, convolutional neural networks, and generative adversarial networks, and variational autoencoder. The full source code is available at https://github.com/GeoX-Lab/ANPyC.
\end{abstract}

\begin{IEEEkeywords}
long-term learning, catastrophic forgetting, adversarial, neural pruning, synaptic consolidation
\end{IEEEkeywords}

%
\IEEEpeerreviewmaketitle

\section{Introduction}
%
%
%
%
\IEEEPARstart{H}{umans} can learn sequential tasks and memorize acquired skills, such as running, biking, and reading, throughout their lifetimes. This ability, namely, continual learning, is crucial to the development of artificial general intelligence\cite{pratama2013panfis}. Existing models lack this ability mainly due to catastrophic forgetting, which means the performance of models in previous tasks degrades when learning upcoming new tasks \cite{McCloskey1989Catastrophic}. What is more, the degradation becomes severe when models integrate a long sequence of tasks. To mitigate this long-term catastrophic forgetting, a straightforward approach is to replay previous data with new data. Works \cite{rebuffi2017icarl,robins1993catastrophic,diaz2014incremental} reinforce historical memories by retrospecting experiences. However, because it takes much memory space to store history data or synthetic data and takes more time to train models, this manner is inefficient for networks with small memory capacity and high model update frequency \cite{Li2017Learning}. Another way to address long-term catastrophic forgetting is dynamic architecture \cite{Rusu2016Progressive,Fernando2017Pathnet:,coop2013ensemble}, which attempts to reserve task-specific resources for individual tasks, such as layers or modules, and to dynamically allocate new resources for upcoming tasks. It is effective but has high model complexity and long search time.

An ideal learning system could sequentially learn multiple tasks without increasing the memory size and computational cost \cite{bargi2017adon} while memorizing the earlier task-relevant knowledge as more as possible, which is still a challenging problem for continual learning. The regularization-based methods  \cite{Li2017Learning} try to preserve the handcrafted model-prior about previous tasks by punishing significant parameters rather than storing parts of previously learned data. For instance, elastic parameter updating methods \cite{Kirkpatrick2016Overcoming,ritter2018online} find a joint distribution of tasks by protecting essential parameters. However, these approaches suffer from poor learning performance when dealing with long sequences of tasks. Although more recent work in \cite{chaudhry2018efficient} simplifies the GEM \cite{Lopez-Paz2017Gradient} and achieves better performance in long-term continual learning, it still requires some additional historical data. Thus, this paper focuses on the regularization-based continual learning approach without additional memory size and computation cost, which are usually needed in memory replay-based approaches.

Mammals can learn hundreds of tasks throughout their lifetimes. One of the most important reasons is the mechanism of memory consolidation based on synaptic plasticity. The transmission efficiency and morphology of synapses, which affects the encoding of memory \cite{gerrow2010synaptic}, are controlled by long-term potentiation (LTP) and long-term depression (LTD). LTP will strengthen the synaptic between two neurons if there is a long-lasting signal between them. It is one of the primary mechanisms that enable learning and memory \cite{bliss1993synaptic}. In contrast, LTD depresses the synaptic connection if the corresponding signal between neurons is weak and intermittent. It serves to selectively weaken specific synapses to coordinate synaptic potentiation caused by LTP \cite{massey2007long}. These two processes are crucial to achieving continual learning since LTP avoids forgetting memory, and LTD avoids lastingly strengthening synapses, which would inhibit the formation of new memory.

In this study, we investigate the issue of long-term catastrophic forgetting by analyzing the two significant challenges of regularization-based continual learning. One of the significant challenges is that the size of the shared parameter-subspace of tasks decreases as the model learns more tasks. Another challenge is the cumulative error among incremental tasks, which origins from two aspects: first, some significant parameters may shift because the regularization is not element-wise punishment; second, the measurement of parameter importance ignores the underlying prior of tasks about the structure of network \cite{courbariaux2016binarized}.

Our works are based on the idea of expanding the shared parameter subspace and decreasing the cumulative error. Inspired by mechanisms of both LTP and LTD, we propose a novel continual learning method, namely, ANPyC, to overcome the long-term catastrophic forgetting issue. In order to achieve LTP, this approach distills the knowledge of the current task into a few parameters, which indirectly expands the common solution space, and it frees up the remaining parameters for learning subsequent tasks. To achieve LTD, we propose a novel parameter-importance measurement that takes into account the structural nature of the network and design a momentum-based weight consolidation policy to protect important parameters element by element. The main contributions of this paper are as follows:

\begin{enumerate}[]
    \item We analyze the underlying reason for the long-term catastrophic forgetting issue in neural network-based continual learning and incorporate neuroscience mechanisms into our long-term learning method. In particular, we propose an adversarial neural pruning and synaptic consolidation approach to tackle this problem.
    \item To precisely protect significant parameters from being forgotten, we design a weight update policy with momentum.
    \item To maintain the structural information of networks, we propose a novel measurement of parameter-importance. This measure utilizes the parameter-connectivity of a neural network. It is a label-free manner, and thus it can generalize to both unsupervised and supervised tasks.
    \item We investigate some regularization methods of overcoming the catastrophic forgetting issue. Experiment results show that our approach is effective and superior to the mainstream regularization based methods for most of the benchmarks and has a strong generalization capability.
\end{enumerate}

\section{RELATED WORKS}
This paper focuses on regularization-based methods, which relate to three areas of work: model pruning based methods, knowledge distillation based methods, and synaptic consolidation based methods.

\subsection{Model Compression Based Methods}
Parameter pruning \cite{smith2018neural,lecun1990optimal,hassibi1993second} are based on the hypothesis that some parameters have little effect on the performance of models after being erased. The fundamental strategy is to search for the optimum parameters that have minimal influence on the loss. PackNet \cite{mallya2018packnet} sequentially compresses multiple tasks into a single model by pruning redundant parameters. The dual memory network \cite{kamra2017deep} is partially based on this idea to overcome catastrophic forgetting by using an external network. These methods still suffer from the long-term forgetting issue. Several improved approaches have been introduced in the literature. For example, combining with synaptic plasticity, \cite{pmlr-v80-serra18a}  narrows the representational overlap between tasks by reducing parameter sharing among tasks. However, it requires additional context-signal of features, and \cite{masse2018alleviating} proposes similar methods by utilizing context-dependent gating as the mask to learn long sequences of tasks. These methods alleviate the problem of long-term forgetting, but they further introduce additional parameters. Compared to these approaches, our method utilizes a greedy pruning mechanism to free low relevant parameters according to network connectivity  \cite{courbariaux2016binarized}; it does not need additional parameters or revision of structure. The empirical study shows that the greedy prune is superior to the context-dependent gating method \cite{masse2018alleviating}  in processing upcoming tasks.

\subsection{Knowledge Distillation Based Methods}
In \cite{hinton2015distilling}, one method based on knowledge distillation packs the knowledge of complex networks into a lightweight network using a teacher-student model, which is also helpful to tackle the problem of catastrophic forgetting. \cite{Li2017Learning} distills the outputs of previous tasks while learning new tasks, but the learning performance would severely degrade when tasks widely vary. \cite{schwarz2018progress} employs an action mechanism to learn the current task and then distills it into the knowledge bag so that it can effectively learn new tasks while protecting previously learned knowledge. \cite{hou2018lifelong} combines distillation with a replay to improve the learning of new tasks, but it reuses a subset of old data to preserve previous tasks. \cite{lee2019overcoming} also proposes a novel distillation loss to overcome forgetting via a large number of unlabeled data. Unlike these works that design explicit distillation regularization, we utilize a greedy prune to distill task-relevant knowledge into a subset of parameters.

\subsection{Synaptic Consolidation Based Methods}
Various synaptic consolidation based methods have been proposed to overcome forgetting via regularization by reducing representational overlap among tasks, such as weights freezing and weight consolidation. Weight freezing, which is inspired by the distributed encoding of human brain neurons, tries to avoid overlaps between crucial functional modules of tasks. For instance, Path-Net
 \cite{Fernando2017Pathnet:} establishes a sizeable neural network and fixes a module of the network to avoid interference from later tasks. This type of method fixes essential parameters of a task to prevent the network from forgetting. However, these methods lack flexibility when facing a long sequence of tasks, and their computation complexity is high.

Weight consolidation tries to identify critical parameters for previous tasks and punish them when training new tasks. A classic method is elastic weight consolidation (EWC) \cite{Kirkpatrick2016Overcoming},  which is inspired by the mechanism of synaptic plasticity. It updates parameters elastically according to parameter importance. Several works \cite{zenke2017continual,aljundi2018memory} propose different parameter-importance measurements, e.g., the sensitivity of parameters to tiny perturbations or label-free. However, none of them have considered the connectivity between the network structure and parameters. \cite{ostapenko2019learning} introduces conditional generative networks to mask layer activation or weights. It is efficient but demands additional parameters. Besides, \cite{chaudhry2018riemannian} improves the distance between parameters in regularization using KL-divergence. The synaptic consolidation based methods encode more tasks with lower network capacity and lower computational complexity than others. Our method proposed in this paper falls into this category, but it is different from existing methods in the following two aspects: first, previous methods are mainly based on the mechanism of LTD and do not specifically address the long-term catastrophic forgetting issue; second, they ignore the cumulative error caused by the regularization and parameter importance measurements.

In comparison, our method aims to alleviate forgetting in long-term learning by integrating the mechanism of LTD to expand the shared parameter subspace of tasks. Meanwhile, we revise the cumulative error by a new parameter-wise updater and structure-aware measurement. We use the updater to control the optimization of parameters rather than a regularization. We propose a novel measurement to mask task-irrelevant parameters rather than randomly mask selectively.

\section{METHODS}
\subsection{Problem Definition }
\begin{figure}[!htb]
\begin{center}
   \includegraphics[width=1.0\linewidth]{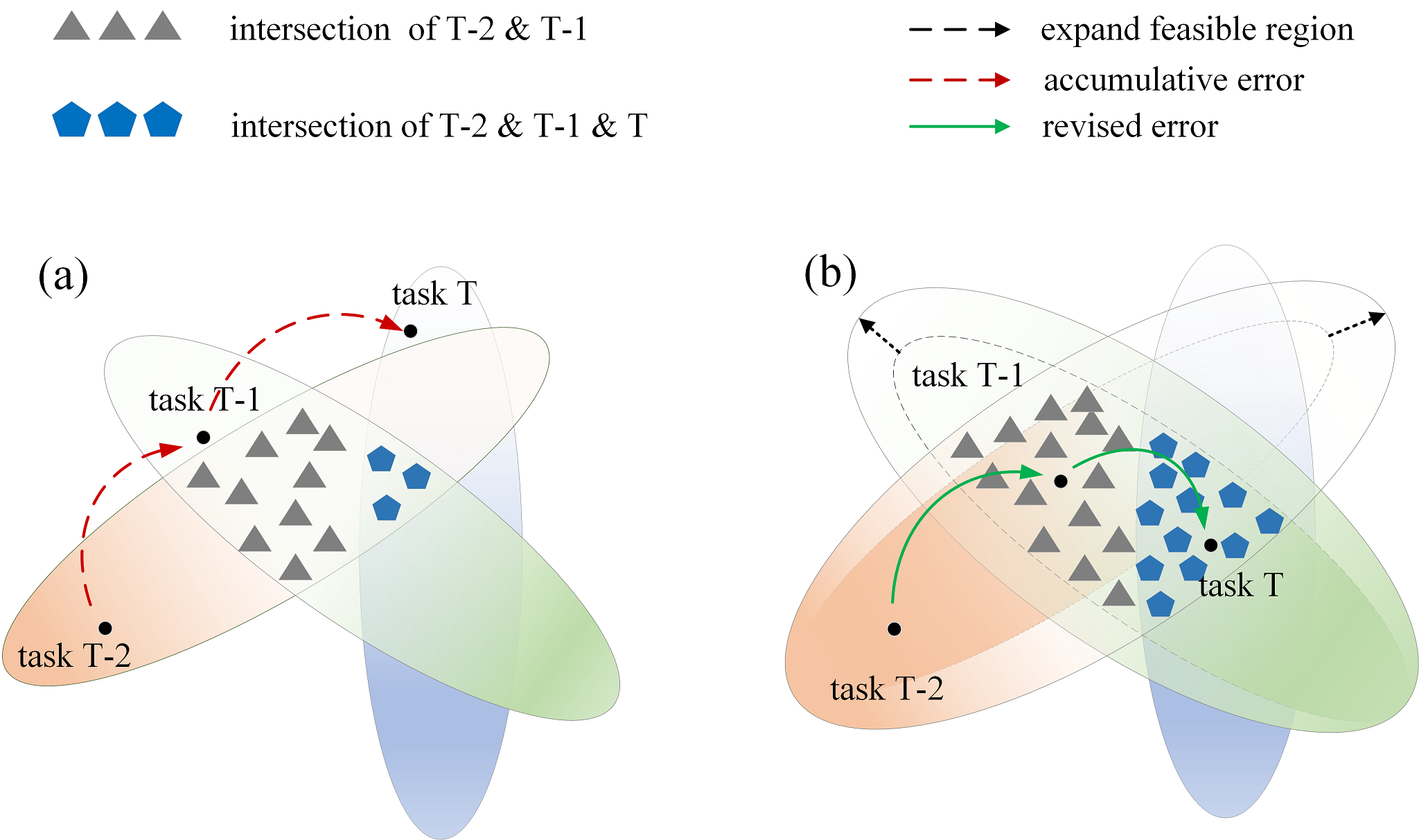}
\end{center}
    \centering
    \caption{(a) The classic process of the consolidation regularizer, e.g., EWC, denoted by the red dashed line. It aims to find a common solution for a sequence of tasks by adapting the parameter space of previous tasks to the current task. The intersection of the parameter subspace (denoted by triangles and pentagons) decreases as the number of tasks increases. In addition, the endpoint may drift away from the right common solution space of the tasks because of the error of the parameter constraints. (b) The black dashed arrow represents feasible region expansion, which is implemented by neural pruning. The neural pruning yields an approximate solution of current tasks using a few parameters. The solid green arrow denotes the revised error. This error revises the consolidation strategy, which is more likely to arrive at the correct common solution space of tasks.}
\label{fig:1}
\end{figure}

Given a sequence of $tasks$ $\{task_{1}, task_{2},..., task_{T} \} $  that are defined by datasets $\{D_{1}, D_{2},..., D_{T}\}$, and a neural network defined by parameters $\Phi$. The objective of continual learning is to sequentially learn all tasks but only the current dataset $D_{T}$ at hand. One current way to address this issue is to find a distribution that fits all tasks from the previous parameter space of tasks, namely,
\begin{equation}\label{eq:1}
    argminf_{\Phi }(D_{1})\overset{D_{2}}{\rightarrow}argminf_{\Phi }(D_{1},D_{2})
\end{equation}
This goal is realized by selectively constraining the optimization trajectory to search for a solution from the common parameter subspace for previous tasks and upcoming tasks. This way is effective with short sequences of tasks. Nevertheless, when it comes to long-term continual learning, it is intractable to search for a solution that satisfies all tasks (Fig. \ref{fig:1}a). The shared parameter subspace is either too small or does not exist when the model learns more tasks, and the cumulative error of the parameter consolidation causes the solution to deviate from the parameter subspace.

To address the above problem, we employ two key strategies in this paper: one strategy is to expand the overlap of the parameter subspace of task (e.g., the region denoted with triangles and pentagons in Fig. \ref{fig:1}b); the other strategy is to use a novel structure-aware synaptic consolidation based updater, which consists of a parameter-wise momentum and a structure-aware parameter importance measure to selectively constrain significant parameters (Fig. \ref{fig:1}b).  In the following, we detail the novel components of the proposed ANPyC, which include the neural pruning, structure-aware synaptic consolidation strategy, and solution to long-term learning.

\subsection{Neural Pruning}
Hence, with the subset of parameters, the learning solution given in Eq. (\ref{eq:1}) can be written as:
 \begin{equation}\label{eq:2}
     argminf_{\Theta }(D_{1})\overset{D_{2}}{\rightarrow}argminf_{\Phi }(D_{1},D_{2})
\end{equation}
where $\Theta\subset\Phi$, is the subset of parameters $\Phi$, and $argminf_{\Theta }(D_{1})$ is the approximate solution of $argminf_{\Phi}(D_{1})$.

Before pruning, we need to answer the question which parameter should be pruned. Here, we utilize the optimal brain surgery \cite{lecun1990optimal} to measure the salience of a parameter. Given that a well-trained model, training parameters \textbf{W} on input \textbf{X} and the error $\textbf{E}=\sum_{i=1}^C p_i \log q_i$. The function of model can be expressed as $f(\textbf{X},\textbf{W})\to \textbf{E}$. If we set the $k$-th parameter $W_k$ to zero, the corresponding change in error $\textbf{E}$ can be expressed as $f(\textbf{X},\textbf{W},W_k)-f(\textbf{X},\textbf{W},0) \to \delta \textbf{E}$. The larger the value of $\delta \textbf{E}$ is, the more important $W_k$ is. Thus it measures the saliency of parameters and its formula of the Taylor expansion can be given as:
\begin{equation}\label{eq:3}
    \Upsilon = (\frac{\partial \textbf{E}}{\partial \textbf{W}})^T\delta \textbf{W} + \frac{1}{2}\delta \textbf{W}^T \textbf{H}\delta \textbf{W} + O(\lVert \delta \textbf{W}\rVert^3)
\end{equation}
Here, $\Upsilon$ is parameter-saliency, $\textbf{H} \equiv \partial ^2 \textbf{E}/\partial \textbf{W}^2$ is the Hessian matrix of the parameters, and $ \partial \textbf{E}/\partial \textbf{W}$ is the first derivative of $\textbf{E}$ with respect to $\textbf{W}$.
The calculation of the Hessian matrix is usually computation-intensive \cite{xu2015optimization}. Here, we introduce the diagonal Fisher information matrix \cite{Pascanu2013Revisiting} to approximate the Hessian matrix. The main advantage is that its computational complexity is linear to the number of dimensions.

We train parameters $ \Phi $ and calculate the salience of parameters $ \Upsilon $. Then, we generate a binary mask in which these salient parameters are set to zeros to prevent them from updating in the current task according to a threshold $ \beta $. These parameters are not actually pruned but remain for later tasks. The objective of pruning is as follow:
\begin{equation}\label{eq:4}
    \Theta = \Phi\left \{ \Upsilon_{i} < \beta \right \}
\end{equation}
where $ \Theta $ is the parameters preserved to encode the current task, and $ \beta $ is a threshold calculated by the prune percentage.

Compared to context-dependent gating \cite{masse2018alleviating} which randomly masks parameters, our method is performed in an iterative training-pruning way rather than pruning these low-saliency parameters once. This greedy pruning approach implicitly distills the previous training phase into fewer parameters during pruning. Thus, it can also be considered as a greedy pruning approach. Besides, our experimental results demonstrate that our approach can perform better than the context-dependent gating [25].

\subsection{Structure-Aware Synaptic Consolidation Based Momentum-Update}

In order to decrease the cumulative error when sequentially learning multiple tasks, we further develop a novel synaptic consolidation strategy for continual learning. It consists of two components: a novel parameter-wise updater which adapts the searching direction for optimization according to the parameter-importance, and a novel parameter-importance measure based on the state of the connection between two neurons.

\begin{figure}[t]
\begin{center}
   \includegraphics[width=1.0\linewidth]{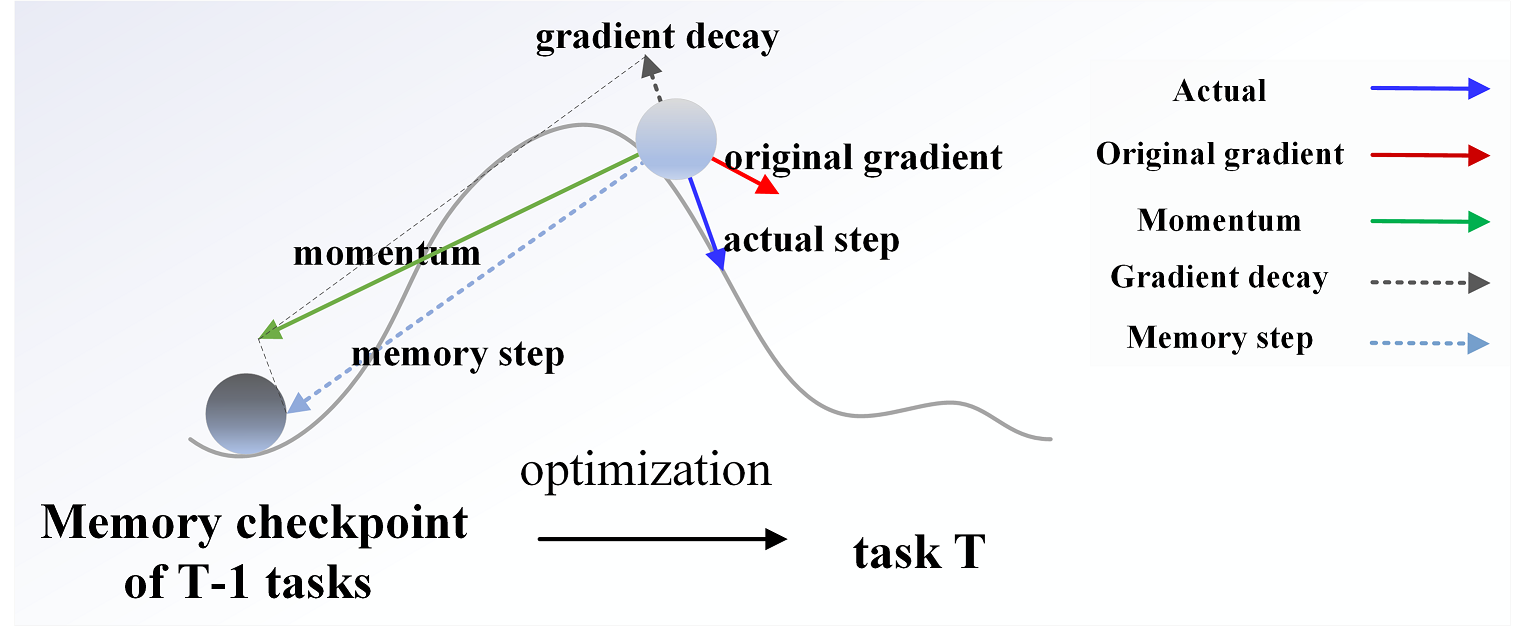}
\end{center}
   \caption{Weight optimization process with momentum. The original gradient, denoted by a solid red line, is revised by the momentum step (the solid green line). The momentum step consists of the gradient decay and the memory step. The direction of the gradient step is opposite to that of the actual step. And the direction of the memory step is toward the memory checkpoint.}
\label{fig:2}
\end{figure}

\subsubsection{Momentum based parameter updating} To ensure that the endpoint of optimization is not far from the previous task when learning a new task, we design a momentum-based updating policy for revising the gradient direction $\bigtriangledown _{\Theta }$ which is calculated via stochastic gradient descent. This policy is implemented as follows:

\begin{equation}\label{eq:5}
    \bigtriangledown _{\Theta } = \bigtriangledown _{\Theta } + Momentum
\end{equation}

As illustrated in Figure \ref{fig:2}, when the optimization point moves toward a new task, which is analogous to a ball rolling up a hill, three forces are related to its movement: the gradient step driven by the target function, which is calculated via classical stochastic gradient descent; the memory step that keeps the ball from leaving the previous memory checkpoint, which ensures the stability of the learning system; and the momentum, whose direction is opposite to the actual step, is the resistance. Here, the memory momentum for the parameter $W_i$ is defined as follows:
\begin{equation}\label{eq:6}
    Momentum = \lambda\left ( -\Omega_{i}W_{i}  +  W_{o}\right )
\end{equation}
where $\lambda$ is a hyper-parameter, of which a large value corresponds to strong momentum, $\Omega_{i}$ is the importance of the parameter $W_{i}$ and $W_{o}$ is the initial value of $W_{i}$ before learning new a task. This momentum defined in Eq. (\ref{eq:6}) can be analogized to the frictional force of the moving ball. Intuitively, we should prevent one parameter with great importance from further changing.

\subsubsection{Measuring the parameter importance through the connectivity of neurons} As stated in Eq. (\ref{eq:3}), the parameter-saliency measures the magnitude of the change in the target function when changing the connectivity state of two neurons. Nevertheless, this measurement implicitly considers the prior knowledge of network structure related to one task.

In addition to the parameter saliency, we perform a further measurement of parameter importance, which usually requires labeled data. To increase the applicability of our method, we avoid the use of labels by utilizing the information entropy to approximate the error of $E$. This is possible because the predicted distribution $q$ approximates the true distribution $p$ on a well-trained model. Besides, the parameter-importance is non-negative. Thus, similar to Eq. (\ref{eq:3}), the parameter-importance is calculated by:
\begin{equation}\label{eq:7}
   \Omega = max(((\frac{\partial H(q)}{\partial \textbf{W}})^T\delta \textbf{W} + \frac{1}{2}\delta \textbf{W}^T \textbf{H}\delta \textbf{W} + O(\lVert \delta \textbf{W}\rVert^3)),0)
\end{equation}
where $H(q)=\sum_{i=1}^C q_i \log q_i$ and $C$ is the number of categories. The strategy is to measure the steady-state of a learning system via information entropy. We explain it as follows: the output distribution of the model will gradually evolve from a random state into a stable state, with decreasing entropy. When the model converges, the system will perform stably on the training data, with low entropy. Therefore, the entropy change is an effective substitute for the loss function change for measuring the steady-state of a learning system.

Compared with EWC and MAS, apart from the fact that our proposed mechanism is unsupervised and can be applied to different learning models, the most important property is that it measures the amount of information about the task in the model structure. This is achieved by calculating the fluctuations in information entropy after removing a connection between certain neurons. It captures the importance of parameters more precisely (details in Part A, Section V).

\subsection{Long-term Learning}
Given a set of $T$ tasks, we calculate the importance  $\Omega_{i,j}^t$ of the parameter $w_{i,j}$ before learning new upcoming $task_{t+1}$, where $i$ and $j$ represent the connections between the $i^{th}$ neuron and the $j^{th}$ neuron, respectively, in a neural network. After learning $task_{t}$, we sum the importances of previous tasks to obtain the following accumulative values:
\begin{equation}\label{eq:8}
    \Omega_{i,j}^{1:t} = \Omega_{i,j}^{1:t-1} + \Omega_{i,j}^{t}
\end{equation}

During learning $task_{t+1}$ on the available dataset $D_{t+1}$, the neural pruning generates a mask $M_{t+1}$ which is calculated by Eq. (\ref{eq:4}). To avoid pruning parameters of previous tasks, the pruning mask of $tasks_{t+1}$ is the intersection of the previous $t+1$ masks:
\begin{equation}\label{eq:9}
    M^{1:t+1}=M^{1:t}\cap M^{t+1}
\end{equation}
Neural pruning distills task-knowledge into a small subset of parameters and releases a large number of non-significant parameters to the current learning task, while synaptic consolidation strengthens parameters of previous tasks. Thus, this adversarial mechanism achieves a balance of preserving old knowledge and learning new knowledge.

We present our algorithm in Algorithm \ref{algrithm-1}.
\begin{algorithm}[h]
\caption{Pseudo Code for ANPyC}
\begin{algorithmic}[1]
\REQUIRE ~~\\
$\textbf{W}^{*}$: old task parameters\\
$\textbf{W}$: new task parameters\\
$X,Y$: training data and ground truth on the new task \\
$tasks$: total number of tasks \\
$H(q)$: information entropy of the output \\
$\textbf{H}$: Hessian matrix \\
$\lambda$: coefficients to control the momentum\\
$\beta$: threshold of salience of parameters for pruning\\
\ENSURE ~~\\
\FOR{each $t \in tasks$}
\STATE $\textbf{W}^{*}$ assign $\textbf{W}$  //Update the old task parameters\\
\STATE $\Omega_{i,j}^{t}=max(0,(\frac{\partial H(q)}{\partial \textbf{W}})^T\delta \textbf{W}+\frac{1}{2}\delta \textbf{W}^T \textbf{H}\delta \textbf{W})$ \\
//Calculate the importance of the parameters of the T-1 tasks\\
\STATE $\Omega^{1:t}=\Omega^{1:t-1}+\Omega^t$ \qquad //Cumulative importance  computation\\
\STATE \textbf{Define:} $\hat{Y}=f(X,\textbf{W})$  \qquad //new task output\\
\STATE $\textbf{W}\leftarrow arg_{\textbf{W}}min(L_{new}(Y,\hat{Y})$ \qquad //Update the new task parameters\\
\STATE $\bigtriangledown _{\Theta } = \bigtriangledown _{\Theta } + Momentum$ \\
\STATE $Momentum = \lambda\left ( -\Omega^{1:t}\textbf{W}  +  \textbf{W}\right )$ //Update the gradients\\
\STATE $\Theta = \Phi\left \{ \Omega > \beta \right \}$ \qquad //generate pruning mask\\
\STATE $ M^{1:t+1}=M^{1:t}\cap M^{t+1}$ \qquad //prune parameters\\
\ENDFOR
\end{algorithmic}
\label{algrithm-1}
\end{algorithm}

\section{Experiments and Results Analysis}
In this paper, we firstly implement the proposed ANPyC on classification tasks with Multilayer Perceptron(MLP) and Convolutional Neural Networks(CNN) in Section IV-B. Then we test the ANPyC on generation tasks with a generative adversarial network (GAN) \cite{goodfellow2014generative} and a Variational Autoencoders (VAE) \cite{kingma2013auto} in Section IV-C. At last, we evaluate the performance of the ANPyC with various pruning strategies and parameter sizes, the contribution of the components of ANPyC, and the influence of two crucial factors (pruning ratio and momentum ratio) in Section IV-D. First of all, we describe the experimental setting in Section IV-A, including dataset, baseline, evaluation metrics, and implementation details.

\subsection{Experimental Setting}

\textbf{Dataset.} For image classification tasks, the permuted MNIST \cite{srivastava2013compete} or split MNIST \cite{lee2017overcoming} is applied to MLP. The CIFAR-10 \cite{Krizhevsky2009Learning}, the NOT-MNIST \cite{Bulatov2011Notmnist}, the SVHN \cite{netzer2011reading}, and the STL-10 \cite{coates2011analysis}, which are RGB images with the size of 32$\times$32 pixels, are used as benchmarks for CNNs. For long-term incremental learning tasks, Caltech-101 \cite{Fei-Fei2006One-shot} is used for the learning with incremental classes. For generative learning tasks, celebA \cite{liu2015deep} and anime face that is crawled from the web are selected as test data, both of which share the same resolution of 96$\times$96 pixels. In addition, MNIST \cite{lee2017overcoming} and SVHN\cite{netzer2011reading} are used for sequentially generating new categories.

\textbf{Baseline.} In image classification tasks, we compare our method with some recent regularization-based methods, including LwF \cite{Li2017Learning}, EWC \cite{Kirkpatrick2016Overcoming}, SI \cite{zenke2017continual}, MAS \cite{aljundi2018memory} and HAT \cite{pmlr-v80-serra18a}. We also compare some traditional methods, including standard SGD with a single output layer (single-headed SGD), SGD with multiple output layers, SGD with frozen intermediate layers (SGD-F), and SGD with fine-tuned intermediate layers (finetuning). Besides, the multitask joint training (Joint) \cite{yuan2012visual} which assumes all task data are available during the training acts as the baseline for evaluating the difficulty of a sequential task. This method provides the ceiling performance since it utilizes the data of all tasks to train the model.

In image generation tasks, DGR \cite{shin2017continual}, which is one of the most effective methods for image generation, is used for performance comparison. DGR is a memory-replay method that requires training additional generative models. In addition, we compare the performance of EWC \cite{Kirkpatrick2016Overcoming} and MAS \cite{aljundi2018memory}, both of which, similar to our ANPyC, are regularization-based approaches without introducing additional parameters at the coding layer for pseudo-data generation.

Furthermore, in our ablation analysis, the mask generation method based on context-dependent gating \cite{masse2018alleviating} is introduced for comparison with our neural pruning method.

\textbf{Evaluation.} We utilize the average accuracy (ACC), forward transfer (FWT), and backward transfer (BWT) \cite{Lopez-Paz2017Gradient} to evaluate the continual learning performance of models: (1) the ACC metric is used for evaluating the average performance over all learned tasks; (2) the FWT metric is used for measuring the suppression of former tasks on later tasks; and (3) the BWT metric is used for measuring the forgetting of previous tasks. As it is more objective to evaluate the difficulty of an individual task by testing the model using multitask joint training \cite{yuan2012visual}, we propose a modified version based on \cite{Lopez-Paz2017Gradient}, i.e., given $T$ tasks, we evaluate the previous $T-1$ tasks after training on the $T$-th tasks. Denoting the accuracy on the $i$-th task for the model trained on the $j$-th task as $P_{j,i}$, and the accuracy on the $j$-th task through the joint learning as $m_{i}$, these three performance metrics can be defined as follows:

\begin{equation}\label{eq:10}
    ACC(i) =\frac{1}{T} \sum_{i=1}^{T}P_{T,i}
\end{equation}

\begin{equation}\label{eq:11}
    FWT =\frac{1}{T-1} \sum_{i=2}^{T}P_{i,i}-m_{i}
\end{equation}

\begin{equation}\label{eq:12}
    BWT =\frac{1}{T-1} \sum_{i=1}^{T-1}P_{T,i}-P_{i,i}
\end{equation}
A higher value of $ACC$ corresponds to superior overall performance, and higher values of $BWT$ and $FWT$ correspond to better trade-off between memorizing previous tasks and learning new ones.

\textbf{Implementation Details.} In the image recognition tasks, we use a multi-head structure, i.e., we configure individual classifiers for each task. The task label is used in the test session, which measures the learning performance with the number of learned tasks rather than in the training session. Specifically, all tasks share the feature learning module. The MLP network is two layers of a fully connected network; for the CNN network, it is VGG-9 \cite{simonyan2014very} and ResNet-18 \cite{he2016deep}. The VGG-9 works for image classification for multiple tasks. The ResNet-18 works for incremental class learning on Caltech-101. We use two types of networks in the image generation tasks, including conditional generative adversarial network (C-GAN) \cite{mirza2014conditional} and VAE. The C-GAN uses an extensible class representation layer to generate different one-hot encodings for each task automatically, while VAE configures specific implicit space layers for different tasks. The convolutional neural networks, including VGG-9 and ResNet-18, are initialized with a pre-trained model in the experiments. While MLP, C-GAN, and VAE use random initialization of network weights. To reduce the interference of random initialization, these networks are trained ten times, and then the mean and standard deviations are calculated to evaluate the final performance of the above methods. We optimize models by the stochastic gradient descent with momentum $\mu$ = 0.9 or $\mu$ = 0.96. The initial learning rate is 0.1, 0.01, or 0.001, and decay with fixed-step = 10 or 20 epoches. We vary the batch size from {128,256} and search the hyper-parameters of our method with grid search. The $\lambda$ ranges from 0 to 10 with step = 1, and the $\beta$ ranges from 0 to 20 with step = 5.


\subsection {Sequential Image Classification}
\subsubsection{MLP \& split MNIST and permuted MNIST}
We implement the proposed method based on an MLP neural network and train it on split MNIST and permuted MNIST. We divide the MNIST into five sub-datasets and train a model with 784-512-256 units. Besides, each classifier has two output units. In Table \ref{tab:1}, we present the experimental results on split MNIST. It shows that Fine-tuning and SGD achieve the best performance in terms of the FWT metric. ANPyC has a similar impact on FWT as Fine-tuning and SGD. HAT has a loss of about 1.5\% on FWT, mainly because it allocates less capacity for upcoming tasks by the mask of hard attention. Our method achieves similar performances on BWT and ACC as HAT, and it only has a reduction of 1.6\% in ACC after learning ten tasks. Also, our method achieves better performance than HAT on FWT. Note that our approach requires no additional parameters.

\begin{table}[t]
\renewcommand{\arraystretch}{1.3}
\begin{center}
\caption{THE RESULTS OF SPLIT-MNIST}
\label{tab:1}
\scriptsize
\begin{tabular}{lccl}
\toprule
Method & FWT(\%) & BWT(\%) & ACC(\%) \\
\hline
SGD & \textbf{-0.32±0.11} & -31.79±0.12 & 62.37±1.29\\
SGD-F & -17.9±0.67& -13.20±0.53 & 84.47±1.85\\
Fine-tuning & \textbf{-0.27±0.13} & -13.20±0.53 & 84.47±1.85\\
EWC & -4.89±0.65 & -6.27±0.49 & 88.90±1.01\\
SI & -5.62±0.51 & -3.55±0.61 & 90.89±0.79\\
MAS & -4.30±0.42 & -2.49±0.19 & 94.42±0.33\\
LwF & -4.45±0.10 & \textbf{-2.13±0.07} & 94.01±0.21\\
HAT & -1.39±0.12 & \textbf{-0.43±0.11} & \textbf{98.65±0.17} \\
Joint & / & / & \textbf{99.80±0.19}\\
Ours & \textbf{-0.45±0.10} & \textbf{-0.78±0.33} & \textbf{98.14 ±1.03}\\
\bottomrule
\end{tabular}
\end{center}
\end{table}

The permuted MNIST datasets randomly permute the pixels of images in the original MNIST dataset. Table \ref{tab:2} presents the results of our approaches and the other nine approaches on ten permuted MNIST tasks. As expected, our method performs best on FWT compared with other regularization based approaches and works as well as SGD and Fine-tuning. SGD-F obtains the highest score on BWT because SGD-F has fixed parameters, which helps protect previous tasks' parameters from being overwritten, but at the cost of reducing ability to learn new tasks. Similar to SGD-F, HAT also has a low BWT because of the use of task-specific masks, but it sacrifices the ability to share features between tasks. ANPyC works not well as MAS and HAT on BWT; however, our method achieves promising learning performances in all three metrics.
\begin{table}[t]
\renewcommand{\arraystretch}{1.3}
\begin{center}
\caption{THE RESULTS ON PERMUTED MNIST}
\label{tab:2}
\scriptsize
\begin{tabular}{lccl}
\toprule
Method & FWT(\%) & BWT(\%) & ACC(\%) \\
\hline
SGD & \textbf{0.92±0.26} & -18.78±0.54 & 70.05±0.69\\
SGD-F & -15.02±0.25 & \textbf{0.15±0.09} & 81.77±0.22\\
Fine-tuning & \textbf{0.77±0.21} & -6.50±0.54 & 80.12±1.42\\
EWC & -1.09±0.19 & -2.59±0.28 & 91.84±0.37\\
SI & -0.71±0.14 & -4.53±0.15 & 90.43±0.48\\
MAS & -1.35±0.34 & \textbf{-1.70±0.32} & \textbf{92.63±0.40}\\
LwF & 0.59±0.18 & -19.36±1.02 & 75.10 ± 1.29\\
HAT & -8.32±0.37 & \textbf{-0.32±0.17} & 89.64±0.39 \\
Joint & / & / & \textbf{95.03±0.19}\\
Ours & \textbf{2.19±0.36} & -3.35±0.37 & \textbf{94.18±0.35}\\
\bottomrule
\end{tabular}
\end{center}
\end{table}

\subsubsection{CNN \& image recognition}
\begin{figure}[t]
\begin{center}
   \includegraphics[width=0.95\linewidth]{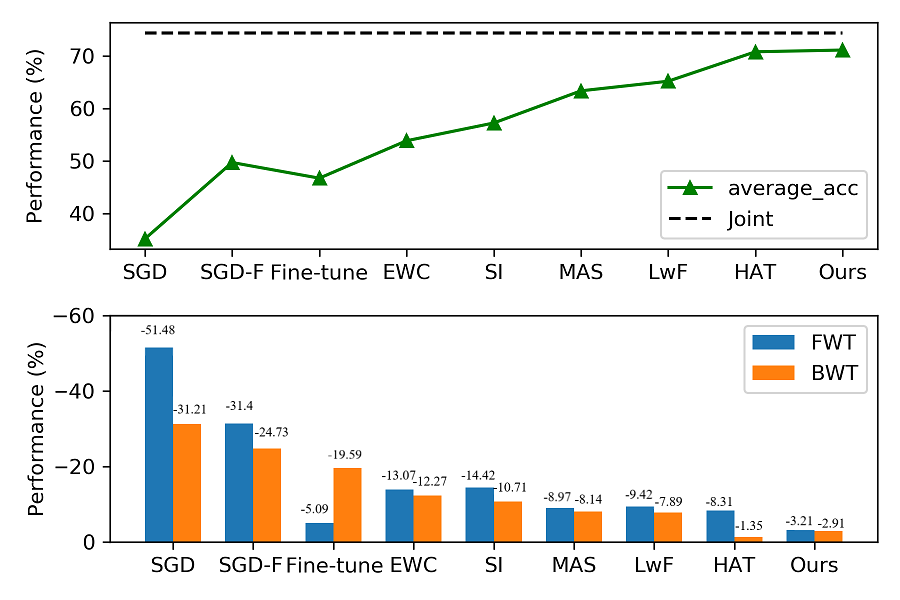}
\end{center}
   \caption{Performance of various methods on sequential image classification tasks. The top shows the results of ACC averaged overall tasks, and the bottom shows the results of BWT and FWT.}
\label{fig:3}
\end{figure}

To verify our method's effectiveness on CNNs, we test continual learning methods in sequential task learning. We use VGG-9 \cite{simonyan2014very}, which consists of 9 convolutional layers and a batch normalization layer to continuously learn MNIST, not NIST, SVHN, STL-10, and CIFAR-10. In our experiments, we resize images as the same size 32$\times$32. We start training a new task only when the current task's loss value no longer decreases in the past three epochs. The result is shown in Figure \ref{fig:3}. ANPyC achieves the best performance in FWT, which is almost one-third of those for LwF and MAS. It shows that our proposed method performs well in alleviating memory dilemma. ANPyC achieves the second-best on BWT, indicating that the network retains the ability to handle previous tasks. In terms of ACC, our method achieves similar performance to HAT but with fewer parameters and computational cost (for details, see Fig. \ref{fig:6}a).

Moreover, Fine-tuning outperforms SGD, which demonstrates that the multi-head classifiers is useful for overcoming forgetting. One possible reason is that the features of tasks at the high layer are highly entangled, and using individual classifiers can slightly alleviate this situation. Overall, ANPyC achieves comparable performance on three indexes, which can effectively relieve the forgetting of a long sequence of tasks.

\begin{figure}[t]
\begin{center}
  \includegraphics[width=1.0\linewidth]{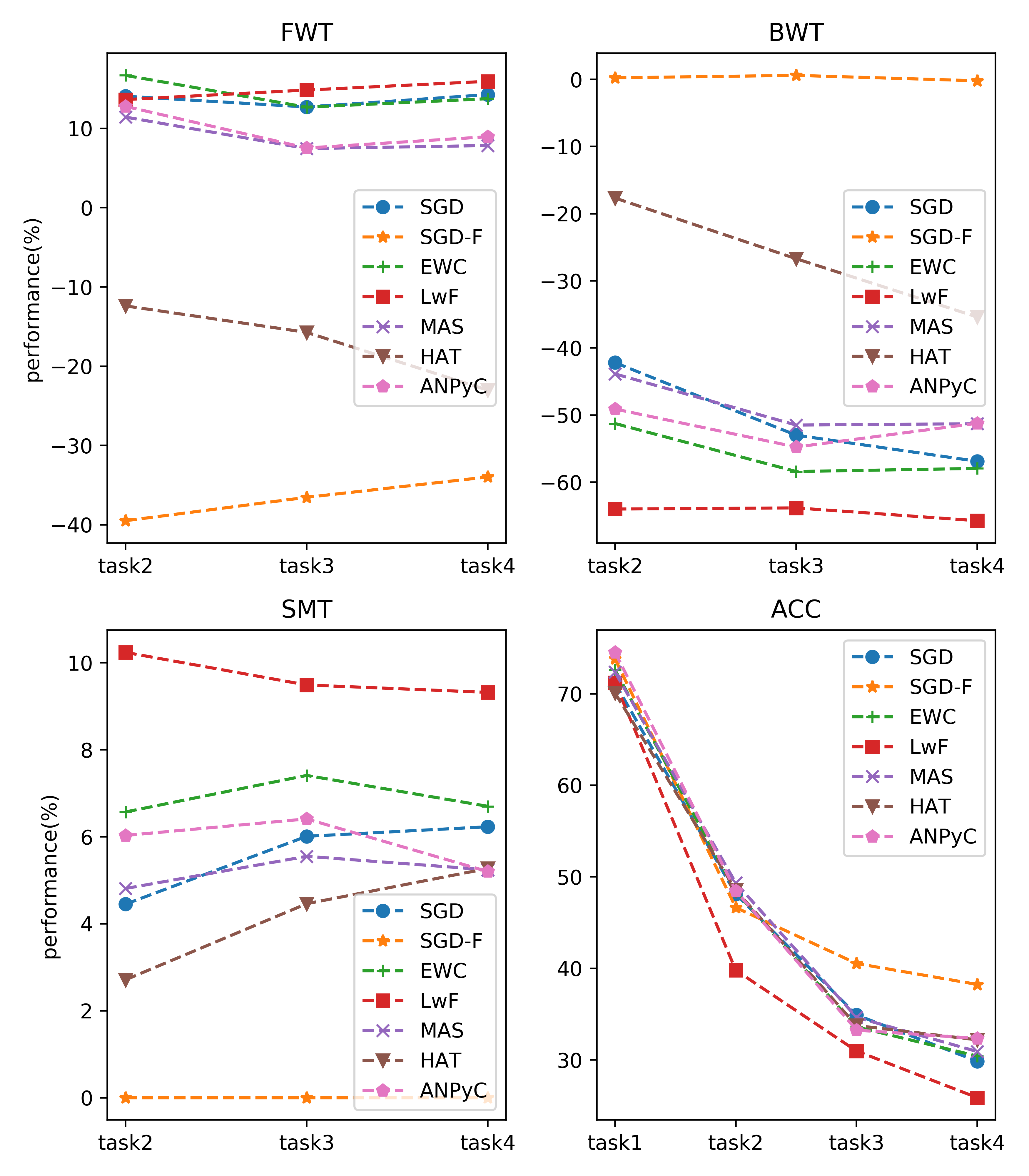}
\end{center}
  \caption{Performance on a subset of Caltech101. The x-axis denotes the tasks that are trained on ResNet-18, and the y-axis denotes the indicators of ACC, SMT, FWT, and BWT. We present the negative values of FWT and BWT in the figures.}
\label{fig:11}
\end{figure}

In addition to sequential domain learning, we use VGG-9 for incrementally learning Caltech-101 (appendix A). The Caltech-101 is uniformly partitioned into four subtasks. As shown in Figure \ref{fig:11}, all current methods do not perform well on large-scale datasets as the number of tasks increases. In the fourth task, the ACC of our approach is less than that of SGD-F. However, it outperforms EWC, MAS, and LwF.  In FWT, our method outperforms all the methods except MAS. In terms of SMT, SGD and MAS outperform our approach.

Nevertheless, our method performs better when learning a more extended sequence of tasks. Notably, it gets the best score on the fourth task in terms of BWT and SMT. Hence, we conclude that ANPyC can preserve the memory of tasks with longer sequences and have higher stability.

\subsection {Incremental Image Generation}
We further incorporate our proposed ANPyC int generative models, including GAN and VAE, on image generation tasks.
For a useful continual learning model, it should generate images that belong to a previously learned task after sequentially learning multiple tasks.

\subsubsection{Continual new category generation with GAN}
In this section, We evaluate the proposed GAN-based ANPyC on MNIST \cite{lecun1998gradient}. The data is divided into ten sub-datasets. Then we select C-GAN as a basic model to sequentially learn how to generate these datasets. This model has a conditional generator that generates specific input vectors for tasks in different domains, thereby mapping various tasks to different spaces. It is critical for continuous generative models because it avoids the signal entanglement caused by the same input signal being decoded into the outputs of multiple tasks, compared with typical GAN models.

\begin{figure}[t]
    \centering
    \includegraphics[width=0.95\linewidth]{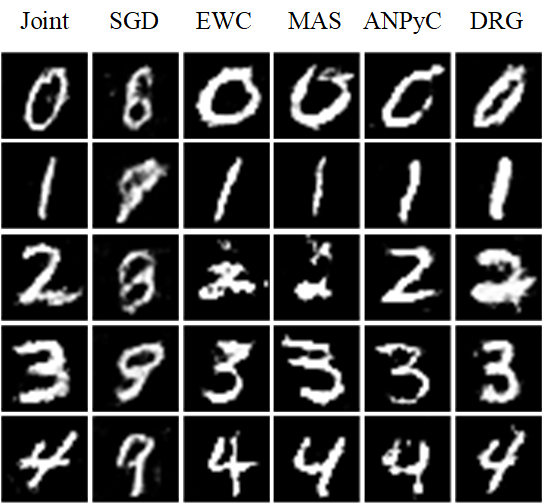}
    \caption {Samples generated by C-GAN with joint training, SGD, EWC, MAS, and ANPyC and DGR on MNIST.}
\label{fig:5}
\end{figure}

To evaluate ANPyC in terms of long-term memory, we test the performance on the previous five tasks individually, using the metric of FID \cite{heusel2017gans}. The FID quantitatively measures the fidelity of the generated data compared to the real images. In Table \ref{tab:3}, Joint is the optimal approach since it utilizes data of all tasks. The methods of SGD, EWC, and MAS, which neither using data from previous tasks nor introducing additional parameters, are used for performance comparison. Besides, we select DGR \cite{shin2017continual} as the suboptimal solution to the continual image generation task. It is a classic rehearsal method that overcomes catastrophic forgetting via replaying a part of previously learned data. Figure \ref{fig:5} presents the results of C-GAN with ANPyC.

Our method is still capable of finely generating the first five digits, as well as Joint and DRG. It means ANPyC works as good as the replay-based method without using additional data in training models. It should be noted that this manner decreases storage and training time. Compared with the above methods, EWC and MAS fail in memorizing the specific digit-2. SGD performs worst. It almost forgets to generate any previous digit. In addition, we quantitatively evaluated these methods in Table \ref{tab:3}. As expected, Joint achieves the best score in FID. The DGR gets a second place as mean = 40.67 because of the accumulated bias in generating history data. ANPyC achieves third place, but it does not utilize historical data. EWC and MAS get a similar score with lower fidelity compared with ANPyC, but still retain some features of previous tasks. SGD fails in memorizing and gets the highest score in terms of FID.

\begin{figure}[t]
    \centering
    \includegraphics[width=0.95\linewidth]{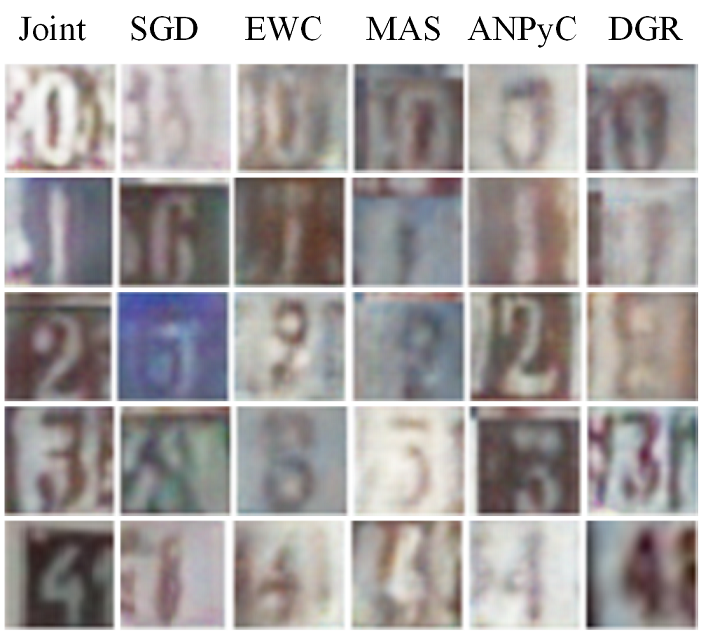}
    \caption {Samples generated by C-GAN with joint training, SGD, EWC, MAS, and ANPyC and DGR on SVHN.}
\label{fig:26}
\end{figure}

To further evaluate the performance of ANPyC, we implement additional experiments on SVHN\cite{netzer2011reading} which is more complex and challenging for image generation than MNIST. We firstly train a C-GAN on digit-0 to digit-4 and then train it on digits 5 to 9. The result is shown in Figure \ref{fig:26}. It illustrates that ANPyC works similarly to Joint and DGR; they all succeed to generate previous digits 0-4, without forgetting. DGR generates most of the previous digits, but it generates some fuzzier images than the former, e.g., the numbers 1 and 2. EWC and MAS are still able to avoid forgetting some numbers, but the level of quality and the number of categories generated are significantly less than the previous three methods. SGD almost forgets all of the previous digits. We calculate the FID in Table \ref{tab:3}. The results support our conclusion. We need to note that ANPyC gets a lower score than DGR, mainly because the SVHN is more challenging to generate, leading to higher accumulated errors in the follow-up tasks. Nevertheless, this does not affect regularization based methods, including ANPyC.

\begin{table*}[t]
\renewcommand{\arraystretch}{1.3}
\begin{center}
\caption{THE RESULTS OF FID IN GAN}
\label{tab:3}
\scriptsize
\begin{tabular}{ccccccc}
\toprule
\textbf{Method} & Joint          & SGD   & EWC   & MAS   & ANPyC & DGR          \\ \hline
\textbf{MNIST}    & \textbf{28.98±0.57} & 68.02±0.82 & 64.70±1.03 & 60.84±1.87 & \textbf{50.71±2.35}  &  \textbf{40.67±1.19} \\
\textbf{SVHN}    & \textbf{60.67±1.39} & 111.71±2.95 & 81.44±2.19 & 73.99±2.77 &  \textbf{67.90±3.11}  &  \textbf{68.10±1.55} \\
\bottomrule
\end{tabular}
\end{center}
\end{table*}

\begin{figure*}[!htbp]
\begin{center}
   \includegraphics[width=0.9\linewidth]{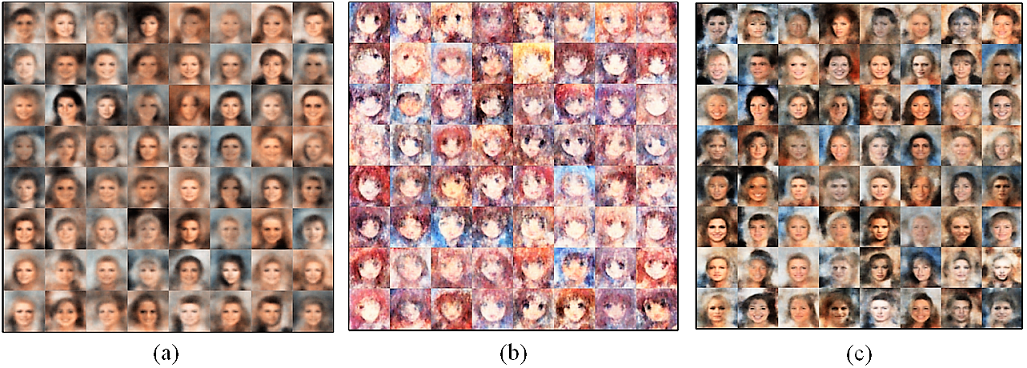}
\end{center}
   \caption{Overcoming catastrophic forgetting from the face dataset to the anime dataset using VAE. (a): The test sample of human faces with a generator from the human face dataset; (b): the test sample of human faces with a generator after training from celebA to the anime face dataset, without using our approach; (c): the test sample of human faces with a generator after training from celebA to the anime face dataset using our approach.}
\label{fig:4}
\end{figure*}

\subsubsection {Overcoming catastrophic forgetting with VAE}	
In addition to GAN, we also verify the effectiveness of our method in VAE models. We train a VAE model from both human faces and anime faces. These two datasets have the same size of 96 $\times$ 96. We set up a VAE with a Conv-Conv-FC encoder layer and an FC-Deconv-Deconv layer on both sides. Then, we use a separate latent variable to train a single task, which is essential because of the significant difference of distributions between these two datasets.

We train models in the following three manners:
\begin{enumerate}[]
    \item Train one model on the Celeba dataset from scratch.
    \item Train one model on the Celeba dataset and subsequently retrain it on the anime face dataset with SGD.
    \item Train the model on the Celeba dataset and subsequently retrain it on the anime face dataset with our proposed ANPyC.
\end{enumerate}
Figure \ref{fig:4} shows samples of human faces produced by the three models. The results demonstrate that our approach can well preserve the skill of human face generation while learning anime faces. The model with ANPyC performs as well as the model trained on the CelebA, whereas the model with SGD loses the ability to remember previously learned human faces. All these results demonstrate that ANPyC has strong generalization in MLP, CNN, GAN, and VAE models.

\begin{figure*}[!htbp]
\begin{center}
   \includegraphics[width=0.9\linewidth]{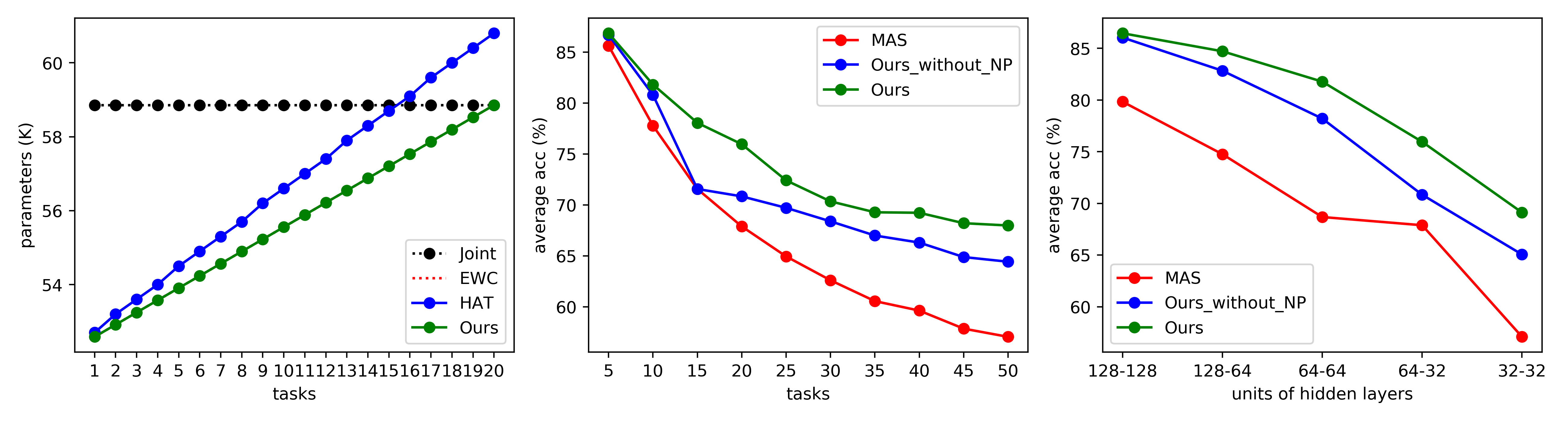}
\end{center}
   \caption{Model size comparison and ablation analysis. (a) The parameter size comparison of methods; (b) The average accuracy for various components combined with the number of tasks that are learned. (c) The average accuracy for various components combination with different model capacities.}
\label{fig:6}
\end{figure*}

\begin{figure*}[!htbp]
\begin{center}
   \includegraphics[width=0.9\linewidth]{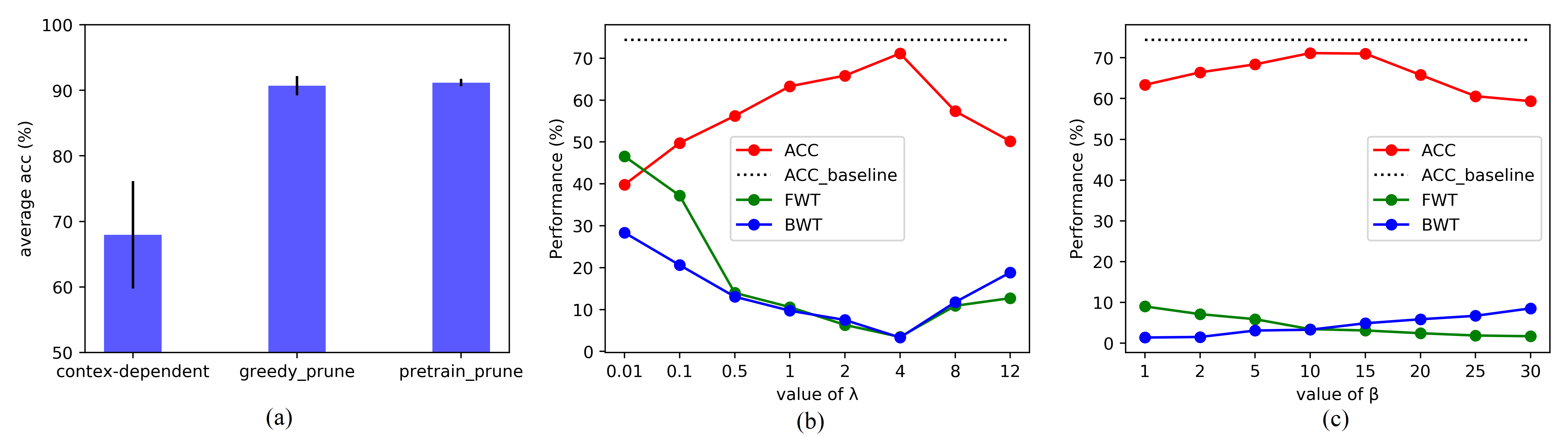}
\end{center}
   \caption{Comparison of pruning strategies and performance of ANPyC under various hyper-parameter values. The dotted black line indicates the baseline of accuracy. (a) average accuracy with various pruning strategies; (b) analysis of $\lambda$ permutation when fixing $\beta$ as 5\% and (c) analysis of $\beta$ permutation when fixing $\lambda$ as 1.}
\label{fig:7}
\end{figure*}

\subsection{Analysis}
To provide an objective assessment, we conduct further experiments to analyze our method's performance by varying the number of model parameters, the neural pruning strategy, the component (ablation study) of ANPyC, and the crucial hyperparameters during the long-term learning process.

\subsubsection{Model parameters comparison}
In Fig. \ref{fig:6}a, we compare the parameter size of models on Permuted-MNIST. Because a separate classifier is configured for each task, HAT, EWC, and our method dynamically increase the output layer's parameters with the number of learned tasks increasing. The parameter growth of the HAT comes not only from the linearly increasing output layer but also from the task encoding vector that increases linearly with the number of tasks. However, the ANPyC and EWC only grow the classifier's parameters, while the backbone network does not add additional parameters. In contrast, the number of classifiers for Joint is initialized to the length of tasks, so it has more parameters in the early session but would not increase with tasks. We can conclude that ANPyC achieves better performance with no additional parameters added to the backbone network.

\subsubsection{Pruning analysis}
To investigate the impact of pruning strategies, we compare the proposed prune strategy with another two approaches: using a hard mask with context-dependent gating \cite{masse2018alleviating} and greedy pruning without pretraining.
Fig. \ref{fig:7}a shows the accuracy of these three pruning strategies on the 20-Permuted MNSIT dataset. The results are averaged over ten runs, each of which randomly selects training and testing datasets. The bar value shows the mean of 10 experiment results, and the black line indicates the standard deviation. It exhibits that the model using the context-dependent gating strategy is unstable (i.e., the highest standard deviation) and performs poorly (i.e., the smallest average accuracy). In comparison, the greedy pruning with or without pretraining gets similar average accuracy, but the former is more stable than the latter. It demonstrates that the pretraining step and greedy pruning are important to achieving stable and high performance.

Besides, we verify the effectiveness of our proposed label-free measurement for parameter-importance. To achieve this, we use an iterative training-pruning to observe the number of parameters of the model. If this parameter importance measure is accurate, only few iterations are producing accurate models. A LeNet is trained on MNIST, and its corresponding results are presented in TABLE \ref{tab:4} (details in APPENDIX B). It shows that only one iteration is required to reduce the parameters by a factor of nearly five without loss of accuracy. Moreover, it can prune even 57 times of parameters in 5 iterations. This indicates our proposed parameter-importance measurement is useful in preserving the most significant knowledge of tasks.

\subsubsection{Ablation analysis}
We conduct an ablation analysis to investigate how each component in ANPyC contributes to the performance. In this experiment, we utilize the average accuracy to evaluate the overall performance of ANPyC. An MLP model is trained on the Permuted MNIST. The pixels of images are randomly permuted in each session of the task. We implement the ablation in two manners: one is to replace the measurement of parameter importance, and the other is to remove the operation of neural pruning. Furthermore, because the length of tasks and the capacity of networks influence the synaptic plastic, which determines the strength of the memory, we control the length of tasks on the fixed networks with a hidden layer of 128-128 units; also, we vary the width of networks to study the contribution of each component in ANPyC.

Fig. \ref{fig:6} shows the corresponding results. It shows that the ACC significantly decreases when we replace our measurement with MAS. The gap between them keeps increasing with the number of learned tasks. Similar results are obtained for networks of different widths. However, we do not find a clear trend for this accuracy gap to increase as the model size gets smaller.
When we removed the neural pruning operation, there is a significant decrease in average accuracy, which occurred both when learning various tasks and different network sizes. In addition to this, as the network size becomes smaller, which means synaptic plasticity is decreased, the difference in performance between our native and the method without neural pruning grows. These results exhibit that the measurement of parameter-importance and the neural pruning is crucial in APNSC, especially in the limited size of networks and long-term learning.

\begin{figure*}[!htbp]
\begin{center}
   \includegraphics[width=0.9\linewidth]{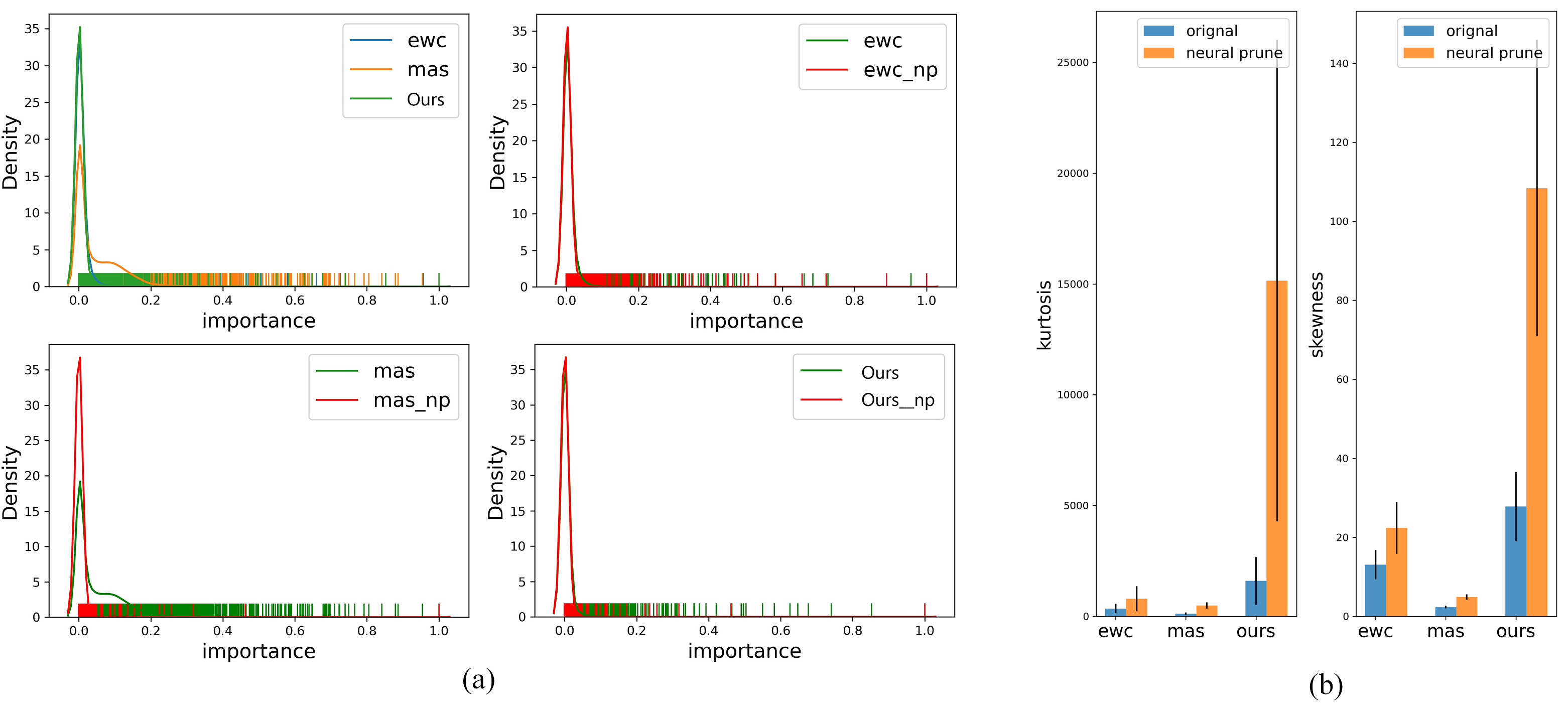}
\end{center}
   \caption{Distributions of the parameter importance values. (a): The distribution of parameter importance. The horizontal axis is the importance value; the vertical axis is the density; the barcodes represent the concentrated area of parameter-importance, the denser the area, the more concentrated the values. Ours is the result without implementing neural pruning; The $Ours_{np}$ is the result with neural pruning. (b) the kurtosis and skewness of parameter importance with various measurements; the black dash line is the standard deviation.}
\label{fig:8}
\end{figure*}

\subsubsection{Hyper-parameters perturbation analysis}
The momentum update and neural pruning are two crucial modules. We analyze the effect of the corresponding hyperparametric changes  $\lambda$ and $\beta$ on the performance of continual learning. The model and data are similar to the subsection of image recognition based on CNN. In Fig. \ref{fig:7}, the results show that our method is robust to hyper-parameter variation in a range of values. According to Fig. \ref{fig:7}b, the network is almost impervious to the resistance of previous tasks when $\lambda$ is 0.01. In this case, the values of all three indicators are extremely poor, and the proposed method and SGD are almost the same at this time. When $\lambda$ reaches 0.1, the proposed method has realized relatively satisfactory performance and has substantially improved on all three indicators. If $\lambda$ is in the range of 0.5 to 4, the performance is relatively stable. The proposed method achieves the best performance with $\lambda=4$. As $\lambda$ continues to rise, the network memorizes too much, which results in a lack of capacity to learn new tasks; hence, the performance on new tasks is lower than the model trained from scratch.

Fig. \ref{fig:7}c shows that models are more robust to the change of $\beta$. The performance fluctuates within 10\% on BWT and FWT when the beta fluctuates in the range of 1 to 30. The FWT decreases with the values grows while the BWT increases with $\beta$ grows, and they achieve a balance when $\beta$ is 10\%. It concludes that smaller $ \beta $ contributes to learning new tasks but losing some memory of old tasks. At the same time, the average accuracy increases with a value of $\beta$ grow, and when it is over 10\%, the average accuracy starts to drop quickly. It reaches an optimum at a $\beta$ value of 15.

\begin{figure*}[htb]
\begin{center}
   \includegraphics[width=0.9\linewidth]{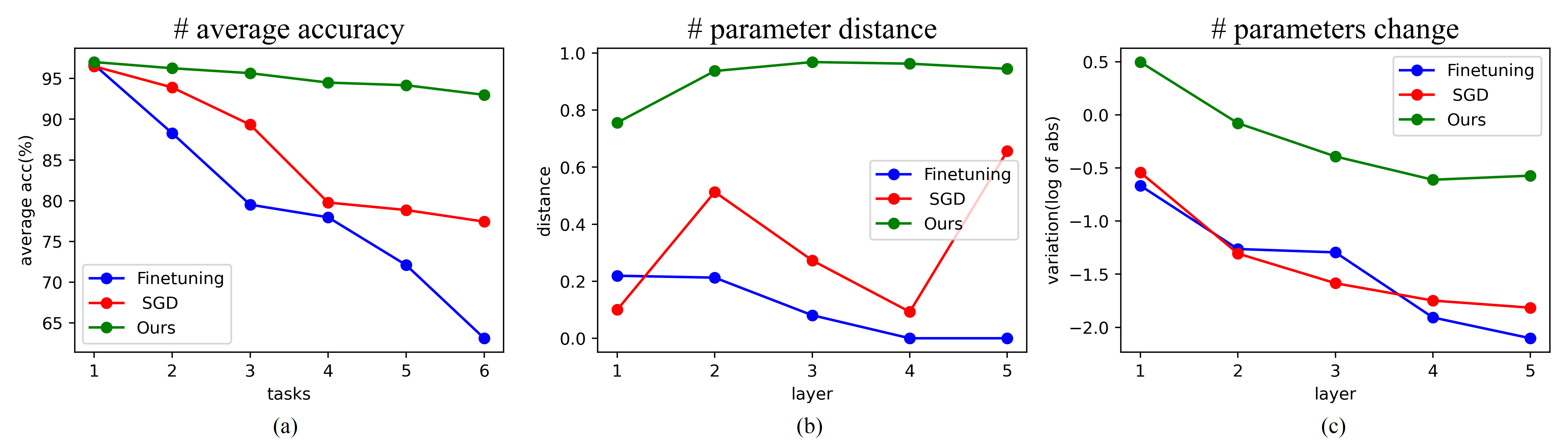}
\end{center}
   \caption{Parameter space similarity and change analysis on Pemuted-MNIST sequential tasks. Each red line corresponds to  standard SGD with a single head, and each blue line corresponds to Fine-tuning and each green line corresponds to ours; (a): Overall average accuracy on 6 permuted MNIST sequential subtasks; (b): similarity of the parameter space; and (c): the parameter variance between parameters of tasks.}
\label{fig:9}
\end{figure*}

\section{DISCUSSION}
\subsection{Synaptic Density Analysis}
An effective continual learning system should balance memory stability and learning plasticity, but maintaining memory while keeping the network scalable is still a crucial question [54]. A typical view is that it is essential to maintain the model's sparsity while stabilizing synapse [25]. Drawing from the above idea, we calculate the distributions of parameter importance values.
It contains two parts: firstly, we compare our proposed measurement with EWC and MAS; and secondly, we validated the effectiveness of the neural pruning mechanism in enhancing sparse synapses on the above three metrics. We achieve it by training an MLP on MNIST in 10 runs. The values of parameter-importance are normalized to ensure these measurements on the same scale. In Fig. \ref{fig:8}a, the results show that our method possesses the nature of the concentrated and polarized distribution of importance values compared with others. The parameters are mainly concentrated at the two ends, which we call polarizability.
Moreover, the vast majority of the parameters are in the low-value region, which we call as concentrate. We also demonstrate that neural pruning can significantly improve the polarizability and concentrated on the parameters' distribution. Specifically, these three measurements benefit from neural pruning, showing an increase of parameters in the low-importance area.

Furthermore, to characterize the distribution of parameter importance values, we used two metrics to evaluate them,  kurtosis and skewness. In Fig. \ref{fig:8}b, our method gets a higher mean kurtosis than other measurements, which holds across standard deviations. A similar phenomenon also exists for the skewness. Besides, the results in kurtosis and skewness consistently show that the measurements with the neural pruning score significantly higher than the naive approaches. Therefore, we conclude that our proposed measure catches significant parameters more precisely, which exhibits lower synaptic density and less forgetfulness. More importantly, neural pruning is an efficient way to promote low synaptic density.

\subsection{Parameter Space Similarity and Changing Analysis}
Interference between old and new tasks in parameter space is the underlying reason why catastrophic forgetting occurs. To verify the effectiveness of ANPyC in maintaining task separability, we analyzed the distance between different tasks in the parameter space. Also, we further analyzed that if our method can maintain a balance between synaptic stability and plasticity, then the parameters can be adjusted to the new task to a greater extent while keeping the old task's solution.

In practice, we sequentially train an MLP model on six tasks with Permuted-MNIST.  The result of the overall average accuracy in Fig. \ref{fig:9}a indicates that our method is more stable and achieves more accurate results as the number of tasks increases compared with SGD and Finetuning.
We utilize the Frechet distance \cite{frechet1906quelques} to measure the similarity of the parameter distributions between the first tasks and the last tasks; see Fig. \ref{fig:9}b. ANPyC allows for more considerable distances between tasks in parameter space, thus avoiding mutual interference. Our method's F value is far greater than those of the other two methods; hence, our approach can effectively control parameter updates according to importance. Moreover, keeping the high-level semantic codes analyzable is more useful to overcome forgetting. Specifically, the F values of ANPyC are greater in deeper layers of the networks. Thus, forgetting occurs mainly in deeper layers.

A weighted sum of the absolute values of differences between the first and the last task is utilized to measure the parameter change. We find that ANPyC allows parameters to be adjusted over a broader range without forgetting. In Fig. \ref{fig:9}c, the result shows that the fluctuation of parameters based on our methods is much more significant compared to other methods.  Thus, it is easy to achieve the trade-off between the intransigence on new tasks and the forgetting on old tasks.
Besides, parameter changes in shallow layers are more extensive. It shows that the consolidation of the constraints on shallow layers is more relaxed than deep ones.  Furthermore, this supports the conclusion in Fig. \ref{fig:9}b.


\section{Conclusions and Future Works}
Long-term catastrophic forgetting limits the application of neural networks in practice. To address this issue, in this paper, we review regularization-based strategies and analyze the causes of long-term forgetting in such approaches: the shrinking of the shared parameter subspace of tasks and the accumulated error of weight consolidation as tasks arrive. We propose the adversarial neural pruning and synaptic consolidation approach to overcome long-term catastrophic forgetting. This approach balances the learning model's short-term and long-term profits by weight pruning and structure-aware synaptic consolidation. The calculation of parameter saliency is similar to optimal brain surgery \cite{hassibi1993second}; however, our method frees parameters from updating and spares them for the following tasks instead of throwing them away. Besides, we assume that the model's structural knowledge is significant and measure it with neuron connectivity. The experimental results demonstrate several advantages of our method:
\begin{enumerate}[]
    \item Efficiency: our approach performs comparably on a variety of datasets for overcoming long-term catastrophic forgetting with no additional parameters and history data storage;
    \item Universality: our parameter-importance measurement considers the structure knowledge underlying network, and it is label-free. Our approach can extend to generative models;
    \item Robustness: our approach has low sensitivity to hyper-parameters.
\end{enumerate}

The evidence suggests that finding an approximate solution of tasks is an effective way to alleviate the memory dilemma. Neural pruning is not the only approach for achieving this solution but provides an idea for combining the neural mechanisms of memory with artificial neural networks; other methods such as knowledge distillation are also feasible. The sparsity of synaptic connection is a significant nature in memorizing for mammals, which means parameter distribution's concentration and polarization in the artificial neural networks. A single strategy is not entirely sufficient for the lifelong learning system. A well-structured constraint for controlling parameter optimization trajectory and well-designed patterns of distributions of parameters may be important pieces of the puzzle to the satisfactory performance of a model in overcoming forgetting. The problem of overcoming catastrophic forgetting remains open, and research on human brain memory provides a potential approach for solving this problem \cite{Hassabis2017Neuro}.

\bibliographystyle{IEEEtran}
\bibliography{Ref.bib}

\newpage
\appendices


\section{Large scale dataset from Caltech-101}

To evaluate our method's performance on a larger dataset, we randomly split the Caltech-101 dataset into four subsets with 30, 25, 25, and 22 classes and divide each part of the subsets into training and validation sets according to the ratio of 7:3. In the experiment, we resize the images to [224,224,3], normalize the pixels into [0,1], and randomly flip the images from left and right to augment the data. We employ ResNet-18 as the backbone. Because the categories of four datasets are not consistent, we add a new separate classifier and a fully connected layer before the classifier for each task. Each new fully connected layer has 2048 neural units, and the dropout rate is set to 0.5. The iteration size and batch size of every task are 100 epochs and 128, respectively. The initial learning rate is set to 0.001, and decay is used every 100 epochs to 90\% of the original. To prevent overfitting, we randomly select the hyper-parameter in the range from 0.5 to 30. Due to the varying numbers of categories in the four subsets, we do not compare our method with SI.

A well-functioning model is expected to be stable under abrupt changes of tasks. To evaluate the stability of the model on unseen tasks, we design an indicator, namely, SMT, as follows:

\begin{equation}\label{eq:13}
    SMT =\frac{ANPyC1}{T-1} \sum_{j=1}^{T-1}D_{j}
\end{equation}
where $D$ is the variance of a single task for sequential learning, which reflects the performance fluctuations of the task.

As the number of tasks increases, none of the above methods perform very well. On the fourth task, the ACC of ANPyC is not as high as SGD-F, but compared with other regularization-based methods, e.g., EWC, MAS, and LWF, our method performs best. On BWT, HAT perform best as the number of tasks increases; our method is second but significantly better than the former on FWT. It is because the HAT uses a hard mask to effectively avoid changes in the parameters associated with old tasks and inhibit these parameters from being used by upcoming tasks. Our method is inferior to SGD and MAS in SMT metrics when the model learns to the second and third tasks. Still, when the model learns to the fourth task, our method are better than all methods except for HAT on BWT, and the same phenomenon occurs in SMT metrics, which indicates that our approach can maintain memory for longer sequences of tasks and has the best stability for different tasks; in the FWT metric, our method is also optimal compared with other methods except MAS which is somewhat better than our approach.


\section{Neural pruning}
We prune the LeNet\cite{lecun1998gradient} on MNIST. The maximum epochs are set to 50, the batch size is set to 100, and the learning rate is set to 0.01. We calculate the importance of parameters after training and prune the insignificant parameters according to the importance threshold. We sequentially conduct this procedure 5 times with various thresholds, and we set the best value as [0.8,0.7,0.5,0.4,0.1]. In Table \ref{tab:4}, the experimental results show that the model compressed with ANPyC balances a high compression ratio and low accuracy loss.

\begin{table}[htb]
\centering
\caption{THE RESULTS OF COMPRESSION ON LeNet}
\label{tab:4}
\scriptsize
\begin{tabular}{m{1cm}<{\centering} m{0.8cm}<{\centering} m{0.8cm}<{\centering} m{0.8cm}<{\centering} m{0.8cm}<{\centering} m{0.8cm}<{\centering} m{0.8cm}<{\centering}}
\toprule
Prune iters      & initial & iter 1  & iter 2  & iter 3  & iter 4  & iter 5  \\
\hline
layer1 W         & 800             & 392     & 271     & 184     & 134     & 129     \\
layer1 b         & 32              & 7       & 3       & 2       & 2       & 2       \\
layer2 W         & 51200           & 27923   & 18582   & 14182   & 11593   & 10220   \\
layer2 b         & 64              & 13      & 4       & 2       & 2       & 2       \\
layer3 W         & 1605632         & 310500  & 87135   & 38756   & 21978   & 19565   \\
layer3 b         & 512             & 100     & 30      & 15      & 9       & 9       \\
layer4 W         & 131072          & 24075   & 6895    & 2674    & 1565    & 1400    \\
layer4 b         & 256             & 51      & 16      & 8       & 5       & 5       \\
total params     & 1789568         & 363061  & 112936  & 55823   & 35288   & 31332   \\
compressed times & /               & 4.93x   & 15.85x  & 32.06x  & 50.71x  & 57.12x  \\
prune ratio      & /               & 79.71\% & 93.69\% & 96.88\% & 98.03\% & 98.25\% \\
test acc         & 98.94\%         & 98.87\% & 98.87\% & 98.83\% & 98.67\% & 98.55\% \\
\bottomrule
\end{tabular}
\end{table}

\ifCLASSOPTIONcaptionsoff
  \newpage
\fi

\end{document}